\pdfoutput=1

\documentclass[11pt]{article}

\usepackage{EMNLP2023}
\usepackage{times}
\usepackage{latexsym}
\usepackage[colorinlistoftodos]{todonotes}

\usepackage{tablefootnote}
\usepackage{graphicx}
\usepackage[printonlyused, nohyperlinks, nolist]{acronym}
\usepackage{lscape}
\usepackage{multirow}

\usepackage[T1]{fontenc}

\usepackage[utf8]{inputenc}

\usepackage{microtype}

\usepackage{inconsolata}

\title{Enhancing Multi-Domain Automatic Short Answer Grading through an Explainable Neuro-Symbolic Pipeline}

\author{Felix Künnecke$^{1}$, Anna Filighera$^{2}$, Colin Leong$^{3}$, Tim Steuer$^{2}$ \\
$^{1}$ Department of Computer Science, Technical University of Darmstadt \\
$^{2}$ Multimedia Communications Lab, Technical University of Darmstadt \\
$^{3}$ University of Dayton Research Institute \\
\texttt{kuenneckefelix@gmail.com}, \texttt{anna.filighera@kom.tu-darmstadt.de}, \\ \texttt{cleong1@udayton.edu}, \texttt{tim.steuer@kom.tu-darmstadt.de} }

\begin{document}
\maketitle
\begin{abstract}
Grading short answer questions automatically with interpretable reasoning behind the grading decision is a challenging goal for current transformer approaches. Justification cue detection, in combination with logical reasoners, has shown a promising direction for neuro-symbolic architectures in \ac{ASAG}. But, one of the main challenges is the requirement of annotated justification cues in the students' responses, which only exist for a few \ac{ASAG} data sets. 
To overcome this challenge, we contribute (1) a weakly supervised annotation procedure for justification cues in \ac{ASAG} datasets, and (2) a neuro-symbolic model for explainable \ac{ASAG} based on justification cues.
Our approach improves upon the \acs{RMSE} by 0.24 to 0.3 compared to the state-of-the-art on the Short Answer Feedback dataset in a bilingual, multi-domain, and multi-question training setup. 
This result shows that our approach provides a promising direction for generating high-quality grades and accompanying explanations for future research in \ac{ASAG} and educational \acs{NLP}.

\end{abstract}

\section{Introduction}
The landscape of educational assessment has changed substantially due to the incorporation of automatic grading systems. In particular, the introduction of transformer models has improved prediction accuracy to the extent that even short-answer questions are graded automatically in commercial systems\footnote{\url{https://new.assistments.org/individual-resource/quick-comments}}. However, such algorithms' decision-making process is opaque, making understanding why a specific grade was assigned extremely difficult for teachers and students. Not only is transparency essential for acceptance and trust, but it is also vital that students comprehend where and why they have made mistakes to foster learning~\cite{shute2008formative,winstone2017supporting,10.3389/fpsyg.2019.03087}. Thus, a simple grade without explanation is insufficient for practical use. 

\begin{figure*}
    \centering
    \includegraphics[width=\textwidth]{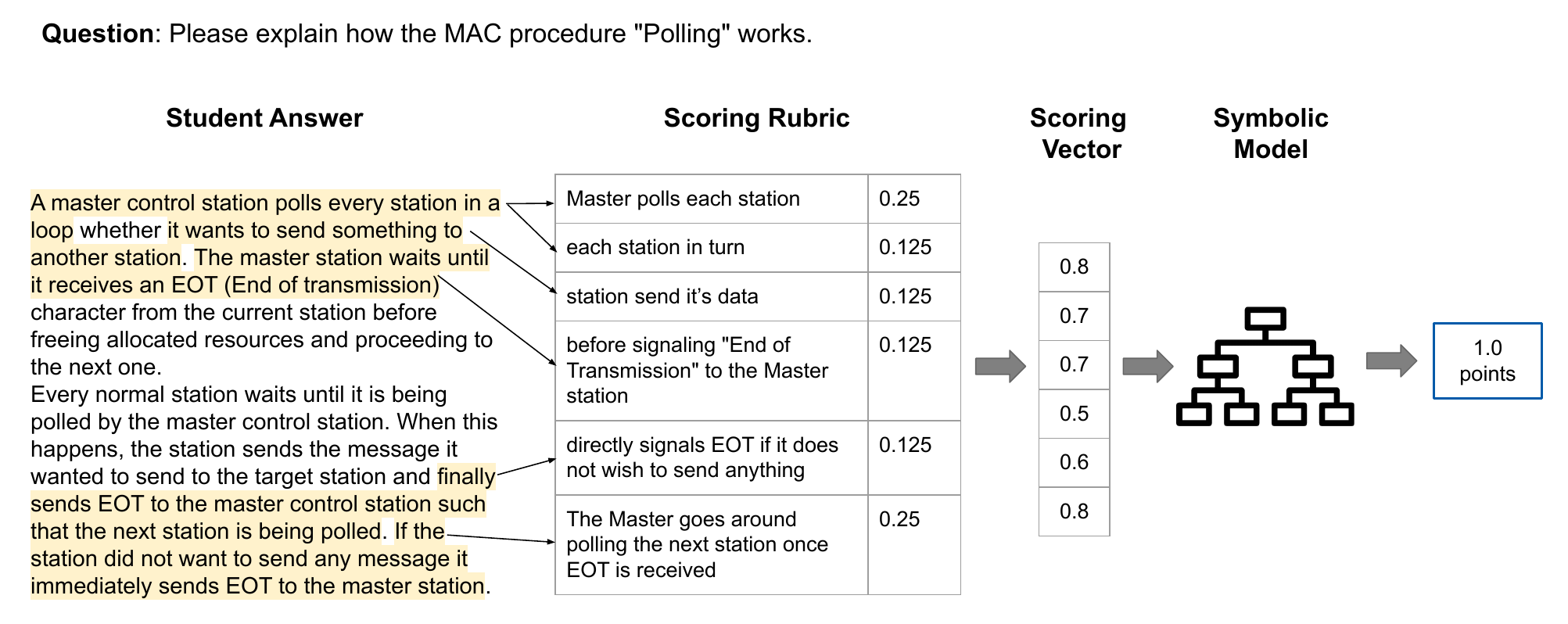}
    \caption{Schematic visualization of our approach with an exemplary student answer. The yellow phrases are the recognized justification cues matched based on their similarity to the scoring rubric. The resulting scoring vector is fed into our symbolic grading models, which are responsible for predicting the final score. This allows our model to provide the actual justification cues, including their similarity to the scoring rubric which was used by the model for the final grading decision.}
    \label{fig:example_approach}
\end{figure*}

Inspired by the justification cue detection task introduced by \citet{Analytic-Score-Prediction-and-Justification-Identification-in-Automated-Short-Answer-Scoring}, we propose a neuro-symbolic pipeline to benefit from the explainability of symbolic models while retaining the flexibility and predictive power of neural networks. The pipeline does not require any specialized annotations beyond what is found in typical \ac{ASAG} datasets, only a scoring rubric detailing which concepts a response should contain and scored responses. First, our approach leverages a weakly supervised transformer to identify important text spans in student responses that contain a concept specified in the scoring rubric, so-called justification cues. Examples of such cues are highlighted in yellow in~\autoref{fig:example_approach}. Then an interpretable symbolic model generates a final grade based on the detected justification cues. In practice, assigned grades can, thus, be explained by which rubric items were identified to what degree in the student's response. 

We demonstrate our pipeline's effectiveness on the bilingual, multi-domain Short Answer Feedback dataset~\cite{Short-Answer-Feedback-Dataset} and make our code publicly available on GitHub\footnote{\url{https://github.com/chefkoch24/neuro-symbolic-asag}.}

\section{Related Work}
Since the research of automatically grading short answers already started in the 90s, the work from \citet{The-eras-and-trends-of-automatic-short-answer-grading} provides a comprehensive overview of the evolution of approaches in the respective domain.
\citet{Survey-on-Automated-Short-Answer-Grading-with-Deep-Learning-from-Word-Embeddings-to-Transformers} further extended this work by incorporating recent developments into their survey and providing a holistic overview from embedding-based methods to transformer-driven approaches and beyond.

However, directly utilizing transformer models to predict grades does not provide explanations for their predictions. Thus, does not foster the necessary understanding and trust in human teachers and learners even if they achieve high accuracy. 
To cover those aspects, \citet{poulton2021explaining} explored various explainability methods for transformer models and evaluated the agreement of important words between the model and human graders. 
In addition, \citet{10.1007/978-981-19-8040-4_5} showed that the best way to support the decision process for human graders considers the visualization of the predicted points together with the matching positions in the student's answer.

To further increase explainability, neuro-symbolic approaches \cite{Kautz_2022} aiming to compensate for the shortcomings of neuronal and symbolic models have gained traction in the \ac{ASAG} community.
Using the representation of scoring rubrics in the form of key items, \citet{Inject-Rubrics-into-Short-Answer-Grading-System} proposed an LSTM model with an attention mechanism on top to inject the given key phrases from the scoring rubric for explicit reasoning. 
Following a human-like task interpretation, \citet{Analytic-Score-Prediction-and-Justification-Identification-in-Automated-Short-Answer-Scoring} proposed to use justification cue detection to identify phrases of the answer that cover the important aspects for the grading and an analytical score prediction, which computes the final score justified with the detected phrases.
For this purpose, they developed a novel dataset in Japanese containing the annotated justification cue spans.
To investigate the effect of the expensive annotations, they experimented on different thresholds of annotated justification cues per question. According to their experiments annotating 100 examples led to human-comparable results for the respective dataset.

Building on this approach, \citet{Automatic-scoring-of-short-answers-using-justification-cues-estimated-by-BERT} utilized BERT on similar experiments and outperformed the baseline with less annotated training data per question.
The main drawback of both approaches is that even in their best-case scenario they rely on manually annotated justification cues, which is an expensive requirement and not given for most existing \ac{ASAG} datasets.

With our approach, we want to overcome the mentioned limitation by following similar task interpretations and utilizing a neuro-symbolic model architecture on a multi-domain, bilingual, and multi-question dataset without manually annotated justification cues.

\section{Approach}
\begin{figure*}
    \centering
    \includegraphics[width=0.8\textwidth]{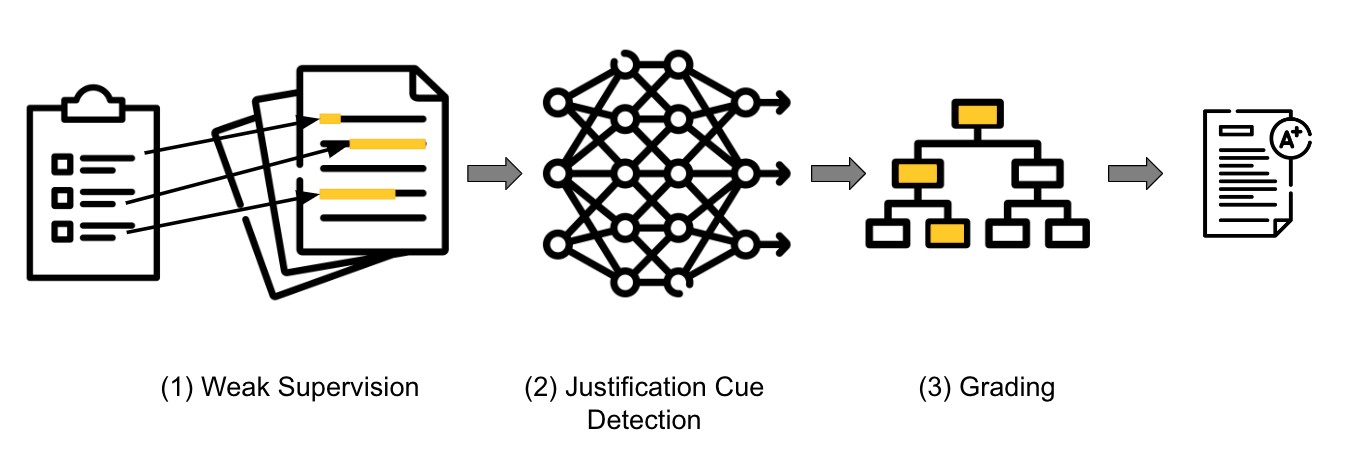}
    \caption{Schematic visualization of the full pipeline of our approach. The pipeline contains three stages: \textbf{(1) Weak Supervision:} annotates the \ac{ASAG} corpus with silver labels. \textbf{(2) Justification Cue Detection:} transformer model trained on the silver labels for finding justification cues in the student answers. \textbf{(3) Grading:} a symbolic model that uses the extracted justification cues for grading based on the similarity to the respective scoring rubric.}
    \label{fig:schematic_pipeline}
\end{figure*}

Inspired by the previously mentioned human-like approach to the \ac{ASAG} task of works from \citet{Analytic-Score-Prediction-and-Justification-Identification-in-Automated-Short-Answer-Scoring} and \citet{Automatic-scoring-of-short-answers-using-justification-cues-estimated-by-BERT}, we propose a multi-step pipeline containing the following stages:
\begin{enumerate}
 \item \textbf{Weak Supervision}: Labeling functions used to supervise the labeling of justification cues in the training data.
 \item \textbf{Justification Cue Detection}: Transformer models for detecting justification cues trained on the weakly supervised data as a token classification or span prediction task.
 \item \textbf{Grading}: Symbolic reasoning in the form of a similarity matching between the detected justification cues and the scoring rubrics by symbolic models to predict the final score for the student answer.
\end{enumerate}

A schematic visualization of the pipeline stages can be found in Figure \ref{fig:schematic_pipeline}.
For our approach, we use analytical scoring rubrics in a structured, tabular format that contain the key elements and corresponding scores for each question.
By retrieving the identified justification cues and mapping their similarity to the corresponding scoring rubric item, along with an interpretable reasoning over it we are able to attain explainability in the grading process.
This enables teachers and students to comprehend the model's decision behind every score. 
A schematic depiction of the full process with an example response is in Figure \ref{fig:example_approach}.
In the following, we proceed to provide a detailed explanation of each step within our designed pipeline.

\begin{figure*}
    \centering
    \includegraphics[width=0.8\textwidth]{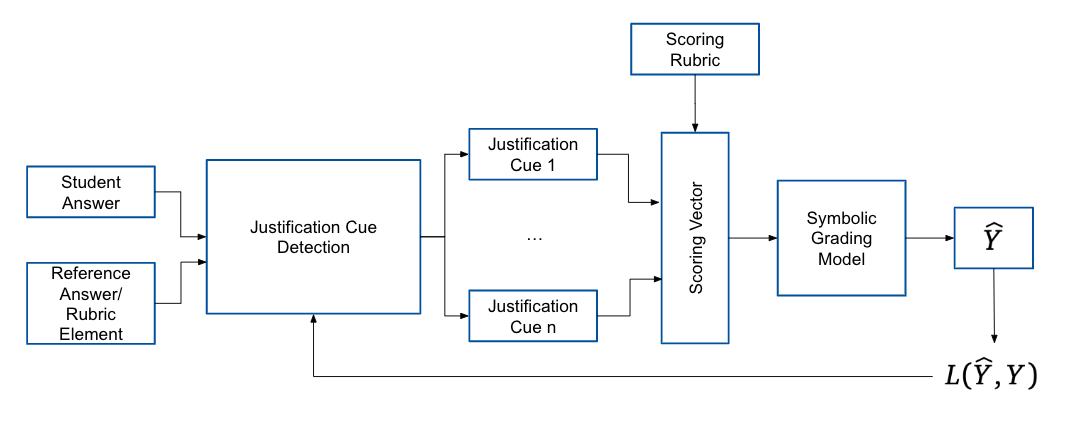}
    \caption{Visualization of the grading process, where the underlying justification cue model retrieves the respective student's answer and context and detects all justification cues. Along the scoring rubric, the justification cues are matched to generate a scoring vector fed into the symbolic grading model. To return feedback on the actual prediction from the grading model, we calculate the loss $L(\hat{Y}, Y)$ and backpropagate it to the justification cue model.}
    \label{fig:final_grading}
\end{figure*}

\subsection{Weak Supervision}
Weak supervision is a strategy to mitigate the challenges of missing annotations in the training data.
Therefore it follows the idea of combining noisy labels, e.g., from crowdsourcing, heuristic rules, and feature-based annotations encapsulated in programmatic functions called labeling functions.
One labeling function alone would be insufficient for annotating the entire corpus.
However, as each function focuses on individual features, their combination provides an appropriate training signal for pre-trained deep learning models \cite{A-Survey-on-Programmatic-Weak-Supervision}.
To implement our weak supervision component, we chose the \textit{skweak} framework from \citet{skweak}. 
As indicated in the study by \citet{Machine-Learning-Approach-for-Automatic-Short-Answer-Grading-A-Systematic-Review}, the majority of features utilized in previous \ac{ASAG} approaches can be categorized into three main groups: lexical, syntactic, and semantic. In line with these findings, we devised two primary classes of labeling functions, namely, \textit{Hard-Matching} and \textit{Soft-Matching}. \\
\textit{Hard-Matching}: uses text features, covering the lexical and syntactic properties in the form of lemmas, word stems, POS-tags, etc. of the justification cue candidates and the scoring rubric item by examining the similarity as a binary decision whether both phrases are equal. 
\\
\textit{Soft-Matching}: on the other hand employs similarity measures like ROUGE \cite{rouge}, BLEU \cite{bleu}, BERTScore \cite{bert-score}, METEOR \cite{meteor}, etc. to assign numerical scores to the justification candidates with regard to the scoring rubric items, where higher scores indicate more similarity between rubric item and candidate and additionally allows to cover semantic similarities.
\\In order to apply our labeling functions effectively, we adopt a two-level approach to identify justification cue candidates within the student answer.
Firstly, we break down the answer into individual sentences and apply the labeling functions iteratively to each sentence.
Secondly, we take the additional step of segmenting these sentences based on their in-sentence punctuation.
This allows us to extract phrases as justification cue candidates, enabling us to handle specific elements such as enumerations as separate entities.\\
Using the output of the labeling functions, we trained a \ac{HMM} over four iterations, which finally outputs probabilities for each token relating to a justification cue. 
These probabilities were then employed as silver labels, resulting in a soft label representation for each token with values ranging between 0 and 1.

\subsection{Justification Cue Detection}
Transformer architectures provide promising results on the common \ac{ASAG} datasets \cite{Investigating-Transformers-for-Automatic-Short-Answer-Grading, Deep-learning-techniques-for-automatic-short-answer-grading}. 
To leverage their capabilities and as one of the main differences to previous works, we utilized them in a question-agnostic and bilingual manner for our justification cue detection.
This is further motivated by \citet{Investigating-Transformers-for-Automatic-Short-Answer-Grading}, finding that multilingual models are suitable for generalizing between languages.
To compare smaller and larger models, we chose \textit{DistilBERT-multilingual-based-cased} \cite{distilBERT} and \textit{mDeBERTaV3} \cite{deberta} as pre-trained transformer models available in the HuggingFace Model Hub. 

Based on our two intuitions, we devised two approaches to identify justification cues within a student's answer.
The first approach involves comparing the student's answer to the reference answer and identifying and marking all justification cues present in the student's response. This comprehensive comparison allows the grader to take into account the context in which the identified cues should be addressed.
The second approach entails finding justification cues through an iterative procedure. Here, the grader examines the list of rubric items and attempts to locate similar phrases within the student's answer. This approach greatly facilitates the clear mapping between the rubric elements and the detected justification cues.
These intuitions can be translated into \ac{NLP} tasks, such as token classification and span prediction, which we will discuss in more detail in the following.\\

\textbf{Token Classification}:
We extended the common token classification problem in two regards.
First, we provide the token classification with an additional context in the form of the reference answer.
Second, we used the probability of each token relating to a justification cue from the weak supervision as a soft label for our token classification model.\\
\textbf{Span Prediction}:
For the span prediction models, we extracted all relevant spans with continuous token labels above a threshold of 0.5 from the weakly supervised corpus. 
Those spans are then iteratively compared against all scoring rubric items using the BERTScore.
The rubric item that achieves the highest score is then used as the silver label for the span prediction model. 
Similar to question-answering models the span predictor retrieves the rubric element and the student's answer aiming to predict the start and end token for the justification cue span.

\subsection{Grading Model}
Based on the weakly supervised justification cue model our grading consists of two components, generating the scoring vector and the final prediction from the symbolic model head. 

For the scoring vector generation, we used our trained justification cue detection model to predict the respective justification cues for the given student answer. 
The detected justification cues are then compared to each scoring rubric item by computing the BERTScore, resulting in question-specific scoring vectors covering the similarity between the scoring rubric and detected justification cues in question-specific scoring vectors.

For our symbolic model heads, we implemented two options. 
Firstly, a naive summation which reasons about the scoring vector given a threshold by summing up all points ascribed to the corresponding rubric items that scored higher than the threshold.
Secondly, we trained question-specific decision trees on the generated scoring vectors.

In this final stage of training our symbolic model heads, we incorporated backpropagation of the loss from the final prediction to our justification cue model. As a result, during this particular step in the pipeline, the justification detection model receives information regarding the correctness of the answer for the first time.
To provide a visual explanation we showcase the full grading process in Figure \ref{fig:final_grading}. 
Thus we used standard mini-batching of 8 samples per batch. 
As the decision tree implementation from scikit-learn does not support batched training we generated all scoring vectors of a training epoch before we trained our final symbolic prediction layer. 

\section{Experiments}
Our experiments aim to compare our neuro-symbolic approach in an optimized pipeline configuration with a purely neural baseline.
Therefore we evaluated our pipeline in each stage on a separate development dataset to find the best configuration.
For our final evaluation, we compared different training corpora and model architectures on the task of scoring automatic short answers.
As we expect that the final model performance highly depends on the justification cue detection, which itself depends on the label quality from the weak supervision, we introduced task-specific metrics for evaluating the intermediate stage of the justification cue detection. These metrics enable a deeper insight into the justification cue detection quality without a labeled gold standard.

\begin{figure}[h!]
    \centering
    \includegraphics[width=0.425\textwidth]{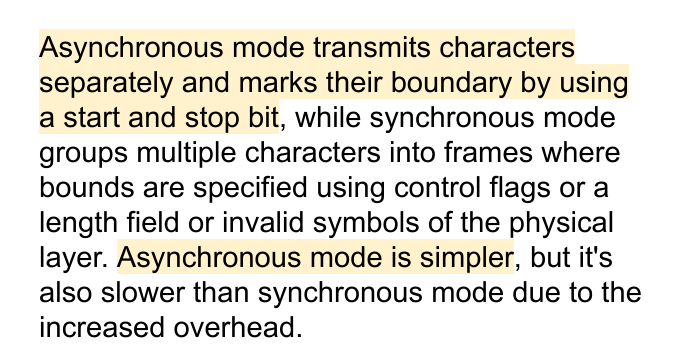}
    \caption{Demonstration of our metrics based on word-tokens for the example from the question: \textit{What is the difference between asynchronous and synchronous transmission mode in the Data Link Layer?}. We highlighted the predicted justification cues in yellow. Number of Justification cues: 2, Average Number of Tokens per Justification Cue: 10, Percentage of Justification Cue Tokens: 0.345.}
    \label{fig:example_metrics}
\end{figure}

\subsection{Task-specific Metrics}
To showcase our task-specific metrics we provide an exemplary student answer in Figure \ref{fig:example_metrics}. \\
\textbf{Number of Justification Cues}: counts the continuous spans in a student's answer, which are labeled as justification cues. 
This metric aims to determine the alignment between the identified justification cues and scoring rubrics motivated by the fact that the scoring rubrics contain multiple elements which should be represented in the student answer. 
In addition, it allows us to assess if the number of justification cues in answers depends on the final grade class, following the assumption that correct answers contain more elements than incorrect ones.
As such, this metric provides a first dimension to understand whether the justification cues align in a way with our scoring rubrics without knowing their actual gold labels.\\
\textbf{Number of Tokens per Justification Cues}: measures the length in tokens of each identified justification cue. This metric enables us to evaluate whether the model captures context within the justification cue. 
This also allows us to determine whether the detection model identifies continuous spans as justification cues instead of labeling individual tokens as is typically done in Named Entity Recognition. 
Since we do not have gold annotated data, we can not determine if it is actually the correct context.\\ 
\textbf{Percentage of Justification Cue Tokens}: refers to the proportion of tokens classified as justification cues, independent of their placement within the student's answer. This is important because models can potentially misidentify entire answers as justification cues, which is undesirable behavior considering that our scoring vector representation requires a matching between the scoring rubric item and justification cue. 
Furthermore, it allows us to assess how much of the answer is grading relevant, following the assumption that correct answers should contain more grading relevant phrases. 

\begin{table*}[h!]
\centering
\begin{tabular}{l|cc|cc||c|c}
\textbf{Task} & \multicolumn{4}{c||}{Token Classification} & \multicolumn{2}{c}{Span Prediction} \\ 
\hline
\textbf{Model} & \multicolumn{2}{c|}{DistilBERT} & \multicolumn{2}{c||}{mDeBERTa} & DistilBERT & mDeBERTa \\ 
\hline
\textbf{\textbf{Context}} & True & False & True & False & - & - \\
\textbf{Macro-F1} & 0.79 & 0.79 & \textbf{0.80} & 0.79 & 0.65 & \textbf{0.67} \\
\textbf{Macro-Precision} & 0.79 & 0.79 & \textbf{0.80} & 0.79 & 0.66 & \textbf{0.69} \\
\textbf{Macro-Recall} & 0.79 & 0.79 & \textbf{0.80} & 0.79 & 0.70 & \textbf{0.73} \\
\textbf{Number of Justification Cues$_{correct}$} & \textbf{2.53} & 2.27 & 2.02 & 2.33 & 0.89 & \textbf{0.95} \\
\textbf{Number of Justification Cues$_{partial}$} & 1.99 & 1.74 & 1.38 & 1.79 & 0.95 & 0.93 \\
\textbf{Number of Justification Cues$_{incorrect}$} & 0.61 & 0.69 & \textbf{0.31} & 0.80 & \textbf{1.00} & \textbf{1.00} \\
\textbf{Percentage of Justification Cues$_{correct}$} & \textbf{0.42} & \textbf{0.42} & 0.38 & 0.39 & 0.27 & 0.27 \\
\textbf{Percentage of Justification Cues$_{partial}$} & 0.35 & 0.37 & 0.30 & 0.34 & 0.30 & 0.31 \\
\textbf{Percentage of Justification Cues$_{incorrect}$} & 0.10 & 0.16 & \textbf{0.08} & 0.20 & \textbf{0.38} & 0.42 \\
\textbf{Token per Justification Cue$_{correct}$} & 11.03 & 12.12 & \textbf{12.55} & 12.50 & 22.06 & \textbf{22.71} \\
\textbf{\textbf{Token per Justification Cue}$_{partial}$} & 9.59 & 10.76 & 11.09 & 10.95 & 18.49 & 18.89 \\
\textbf{\textbf{Token per Justification Cue}$_{incorrect}$} & \textbf{7.00} & 8.42 & 10.46 & 11.19 & \textbf{16.56} & 17.22 \\
\end{tabular}
\caption{Results for the neural justification cue detection on the development dataset. The task-specific metrics are reported for the final class (\textit{correct, partial, incorrect}), averaged over the entire development dataset. We have highlighted the best results, determined based on our prior assumptions, to emphasize the respective performances from the models.}
\label{tab:results_justification_cue_detection}
\end{table*}

\subsection{Dataset}
We used the \ac{SAF} dataset from \citet{Short-Answer-Feedback-Dataset}. 
The dataset contains a bilingual corpus of German and English student answers from different domains and questions.
For the evaluation of the model's performance, it contains two test sets, one for testing a model's capabilities on new answers to questions it was trained for (unseen answers) and one for entirely novel questions (unseen questions). 
Since our symbolic decision tree heads are question-specific, we focus in this work on the unseen answers test split. 
Thus, we experiment with six German questions from a micro-job training and 26 English questions from a communication networks university lecture.
Furthermore, using this dataset allows us to access the raw annotation guidelines and scoring rubrics for each question to extract our scoring rubric representation in a manual procedure from the raw data. The extracted scoring rubrics are published in our GitHub repository.

\subsection{Neural Baseline}
The current baseline on the \ac{SAF} dataset is set by monolingual fine-tuned mBART \cite{mbart} models for the short-answer feedback generation. 
However, the generated output contains a separate verification feedback label which can be used for evaluating the \ac{ASAG} performance \cite{Short-Answer-Feedback-Dataset}.

\subsection{Justification Cue Detection Results}
To determine the justification cue detection model for our final pipeline configuration, we compared the performance of DistilBERT and mDeBERTa on the token classification and span prediction task on the development set before evaluating the final \ac{ASAG} performance on the test set.\\
\textbf{Token Classification}:
The results in Table \ref{tab:results_justification_cue_detection} showed that the weakly-supervised metrics for token classification were generally within the same range for both model variants, with slightly better performances achieved by the variant containing the reference answer as context. 
The task-specific metrics showed more diverging results compared to the standard token classification metrics. 
Revealing that the average number of justification cues for all four models achieved the highest value for correct answers compared to partial correct and incorrect answers, which follows our underlying assumptions.
Interestingly, the DistilBERT models detect more justification cues in the correct answers compared to the mDeBRTA model variants. 
The most apparent finding was that the models showed generally similar performances, independent of whether a context was provided. 
This is an unexpected and counter-intuitive finding as we expected that the context benefits the justification cue detection, similar to previous works in the \ac{ASAG} domain.
\\
\\
\textbf{Span Prediction}:
For the weakly-supervised span prediction metrics, the mDeBERTa model outperformed the DistilBERT. 
However, the span prediction models tended to annotate longer justification cues compared to the token classification models. 
This is particularly interesting as they were trained to find exactly one justification cue element given the rubric element, which makes the metrics only comparable within the same task interpretation.
A potential reason for the prediction of longer justification cues is that the model may maximize the justification cue density in the detected span, which potentially results in blurry justification cue boundaries.
We attribute this behavior to come most likely from unsupervised training data generation for the span prediction architecture, which is further indicated by the fact that both models learn to detect all the annotated justification cues in incorrect answers.\\

After comparing both model architectures, we did not decide on one task interpretation and continued to compare both on the final task. 
However, to keep comparability between both architectures, we ran all our final experiments on mDeBERTa architecture and chose the context-aware token classification model for our final scoring task experiments.

\subsection{Scoring Setup}
For our final experiments of scoring short answers automatically, we compared three training data configurations and four model architectures. 
Based on the assumption that additional training examples might benefit the model's performance.
We chose the following data configurations:
\begin{enumerate}
\item \textit{Monolingual}: the training data contains only answers to German or English questions.
\item \textit{Bilingual}: the training data contains answers to German and English questions.
\item \textit{Unseen}: the training data contains additional bilingual training examples from German and English answers and questions extracted from the unseen question evaluation set.
\end{enumerate}

In addition, we compared our underlying token classification and span prediction models with different symbolic model heads such as naive summation and decision trees for the final scoring results.

\subsection{Scoring Results}
We present our results for scoring short answers automatically in Table \ref{tab:final_results_scoring}. 
The results show that our models outperformed the baseline on English and German data with all model variants using a \ac{DT} as the final symbolic model head. 
We achieved the best results with the SP + DT$_{bilingual}$ by lowering the \ac{RMSE} baseline by 0.244 for the German data and 0.3 for the English data. 
Moreover, we observe that the summation model heads yield insignificant results overall.
Comparing the results from the different data splits, we observe that the incorporation of unseen answers into training has no positive impact on span prediction but improves the token classification compared to the bilingual corpus with only the seen answers.
In addition, we found that the \ac{RMSE} for the German is in all configurations slightly higher compared to the English.

\begin{table}[h!]
\centering
\begin{tabular}{lcc}
\textbf{Model}         & \multicolumn{2}{c}{\textbf{RMSE}}  \\
                       & DE             & EN                \\ 
\hline
mBART Baseline\tablefootnote{German baseline: \url{https://huggingface.co/Short-Answer-Feedback/mbart-score-finetuned-saf-micro-job} \\ English baseline: \url{https://huggingface.co/Short-Answer-Feedback/bart-score-finetuned-saf-communication-networks} [Accessed 30.04.2023]}         & 0.333          & 0.373             \\ 
\hline
TC + DT$_{unseen}$       & 0.093          & 0.076             \\
SP + DT$_{unseen}$       & 0.096          & 0.087             \\
TC + S$_{unseen}$        & 0.428          & 0.536             \\
SP + S$_{unseen}$        & 0.428          & 0.639             \\ 
\hline
TC + DT$_{monolingual}$  & 0.197          & 0.083             \\
SP + DT$_{monolingual}$  & 0.108          & 0.084             \\
TC + S$_{monolingual}$   & 0.433          & 0.555             \\
SP + S$_{monolingual}$   & 0.424          & 0.562             \\ 
\hline
TC + DT$_{bilingual}$ & 0.166          & 0.094             \\
SP + DT$_{bilingual}$ & \textbf{0.089} & \textbf{0.073}    \\
TC + S$_{bilingual}$  & 0.448          & 0.587             \\
SP + S$_{bilingual}$  & 0.432          & 0.578            
\end{tabular}
\caption{Final results for scoring on the test set of unseen answers. TC = Token Classification, SP = Span Prediction, S = Summation, DT = Decision Tree, \textit{unseen} = Training data incl. unseen questions, \textit{monolingual} = Training data only contains the respective language, \textit{bilingual} = Training data contains the full corpus excl. unseen questions.}
\label{tab:final_results_scoring}
\end{table}

\section{Analysis}

As our approach not only strives for decreasing automatic metrics on \ac{ASAG} datasets, we analyzed the explainability features from the justification cues and the scoring vectors for our results further.

\subsection{Token Classification vs. Span Prediction}
To compare the explainability of the predictions between our two task interpretations, we analyzed the justification cues from our best performers in more depth. 
We found 217 cases where the model did not detect justification cues, so the scoring vectors contained only zeros for the token classification model.
Comparably, we did not observe this behavior in the same order of magnitude for the span prediction model, as we only found four cases with no justification cues.

Furthermore, we discovered that our span prediction model tends to predict duplicated spans as justification cues. On average, we observed 1.78 duplicated spans per answer.
To our knowledge, this is mainly due to the unsupervised generation of training data. 
Overcoming this behavior would be possible by enhancing the labeling procedure by including the BERTScore as span soft labels, similar to the training of the justification cue detection based on token classification.\\ 
\subsection{Question-specific performance}
We further conducted additional analysis to examine the performance of our models on a question-specific level.
Therefore we interpreted the predicted scores from 0 to 1 as a 9-class problem and looked at the classification metrics like Accuracy, Macro-F1, and Weighted F1 score to gain deeper insights into our best-performing model. 
Our analysis showed diverging results, as for some questions like 5.12 and 6.1\_IPP, we achieved weighted F1 scores of 1.0, whereas for other questions like  10.2\_TC or 6, only 0.17 and 0.03. All results are available in Appendix \ref{appendix:question-specific-results}.
As we aimed to understand those differences we investigated several potential reasons.\\
\textbf{Rubric Length}:
One reason for the different performances on the question level could be the scoring rubric length underlying the assumption that long lists of rubric items are more difficult to answer.
To determine the correlation between the rubric length and the question-specific performance, we calculated the Pearson correlation coefficient between the rubric length and the question-specific metrics, which led to a correlation between length and Weighted-F1 of 0.22, with a p-value of 0.22, a correlation of 0.35 for the Macro-F1 with a p-value of 0.04, and for the accuracy a correlation of 0.21 with a p-value of 0.26.
According to those results, it indicates only a small positive correlation with the Macro-F1. 
Nevertheless, more was needed to fully explain the observed behavior. \\ 
\textbf{Question-Type}: 
In addition, we analyzed the question types that led to the differences. 
Therefore we looked exemplary at our good-performing questions, which are: 
6.1\_IPP: "\textit{What are the objectives of IPv6? Please state at least 4 objectives.}" or 5.12: "\textit{Discuss 3 methods (each with at least one advantage and disadvantage) that address the problem of duplicate packets on the transport layer in a connection-oriented service.}".
Both questions expected an enumeration of differences by the student.
Surprisingly, the low-performing question 10.1\_TC: "\textit{State at least 4 of the differences shown in the lecture between the UDP and TCP headers.}" underlies the same question pattern and is answerable with an enumerated list from the student.\\ 
\textbf{Label Distribution}:
We further investigated two potential factors that could explain our results regarding the data itself.
Firstly, it is evident that for the final question-specific grading we have significantly less training data compared to the question-agnostic justification cue detection, as the underlying dataset comprises 32 different questions.
Secondly, analyzing the label distributions exhibits imbalances for most of the questions. 
In certain cases, there are even questions where no incorrect samples are available for training but occur during testing.
Based on these findings, we conclude that the variances in label imbalances and the constrained problem space are the likely reasons for the varying performances observed across different questions.

\section{Conclusion \& Future Work}
Motivated by the lack of explainability in current \ac{ASAG} systems, we proposed a neuro-symbolic architecture by splitting the task into the neural justification cue detection, which detects phrases in the student answer that justifies the final grading decision and symbolic reasoning over those detected justification cues by incorporating a similarity matching to the scoring rubrics. Besides, we overcame the challenge that our dataset did not contain justification cue annotations by utilizing weak supervision.
Drawing from the analysis of our results, we observed the most promising results from the span prediction model. Despite we see the potential for improvement by using soft labels similar to our experiments with the token classification. In addition, we recommend incorporating the supervised class label from \ac{ASAG} task in the justification cue detection. This could be achieved similarly to the work from  \citet{Automatic-Patient-Note-Assessment-without-Strong-Supervision} using a multi-task learning setup and predicting the grade and justification cue at the same time. Furthermore, our analysis showed that the scoring vector representation for grading is not always sufficient as it requires domain knowledge in particular for partially correct answers.
To conclude our work, we showed that explicit justification cue detection is feasible to improve explainability in \ac{ASAG}. Furthermore, we believe that detecting justification cues can benefit other educational \ac{NLP} domains, such as short answer feedback generation. To further investigate justification cue detection in the educational \ac{NLP} research, we think that human-in-the-loop approaches utilizing the justification model as a grading assistance would allow the creation of an annotated justification cue datasets and the measurement of the impact of such a system on the final grading process.

\section*{Limitations}
In this paper, we provide a neuro-symbolic approach for \ac{ASAG}. However, our approach has the following limitations.
We only evaluate our approach to unseen answers of the SAF dataset, which relates to our implementation of question-specific symbolic models as decision trees, that did not support zero-shot learning techniques to evaluate unseen questions. 
Furthermore, we manually extracted the scoring rubric elements from the reference answer and did not align these with the original question authors. 
These limitations mean that we cannot use the method for other benchmark datasets without further ado. 
Due to a lack of gold-labeled justification cues, we did not implement a full hyperparameter search to determine the soft-matching thresholds for our weak supervision component. Instead, we used our best guesses from the prior analysis of the individual labeling functions using our introduced task-specific metrics. 
The code in our implementation is not fully optimized for performance purposes and may not always contain the most effective implementations. But it represents the idea and is, to our knowledge, bug-free during testing in the development phase.

\section*{Acknowledgements}
We want to thank all the reviewers for their attentive and insightful help and feedback. In addition, we would like to thank the Multimedia Communications Lab at the Technical University of Darmstadt for providing the compute resources to conduct this research.

\begin{acronym}
\acro{ASAG}{Automatic Short Answer Grading}
\acro{NLP}{Natural Language Processing}
\acro{QWK}{Quadratic Weighted Kappa}
\acro{RMSE}{Root Mean Squared Error}
\acro{SAF}{Short Answer Feedback}
\acro{DT}{Decision Tree}
\acro{HMM}{Hidden Markov Model}

\end{acronym}

\bibliography{anthology,custom}
\bibliographystyle{acl_natbib}

\clearpage
\appendix

\section{Appendix}
\label{sec:appendix}

\subsection{Weak Supervision Labeling Functions}
We provide a table with all of our designed labeling functions including their feature level, matching class, and the detail level of the justification cue candidates.
\begin{table}[ht]
\centering
\begin{tabular}{lllc}
\textbf{Labeling function}                           & \textbf{Feature level}                & \textbf{Candidates}     & \textbf{Matching} \\ \hline
Noun Phrase match                 & lexical, syntactic           & Sentence         & Hard     \\ 
Lemma match                       & lexical                      & Sentence         & Hard     \\ 
POS match                         & syntactic                    & Sentence         & Hard     \\ 
Shape match                       & lexical, syntactic, semantic & Sentence         & Hard     \\ 
Stem match                        & lexical                      & Sentence         & Hard     \\ 
Dependency match                  & syntactic                    & Sentence         & Hard     \\ 
Lemma match without stopwords     & lexical                      & Sentence         & Hard     \\ 
Stem match without stopwords      & lexical                      & Sentence         & Hard     \\ 
POS match without stopwords       & syntactic                    & Sentence         & Hard     \\ 
Dependency match without stopwords & syntactic                    & Sentence         & Hard     \\ 
N-gram overlap (1 - 5)            & lexical                      & Phrase           & Soft     \\ 
ROUGE (1 - 5)                     & lexical                      & Phrase           & Soft     \\ 
ROUGE L                           & lexical                      & Phrase, Sentence & Soft     \\ 
Word Alignment                    & lexical, semantic            & Sentence         & Hard     \\ 
BERTScore                         & semantic                     & Phrase, Sentence & Soft     \\ 
BLEU                              & lexical                      & Phrase, Sentence & Soft     \\ 
METEOR                            & lexical, semantic            & Phrase, Sentence & Soft     \\ 
Jaccard Similarity                & lexical                      & Phrase           & Soft     \\ 
Jaccard Similarity lemmatized     & lexical                      & Phrase           & Soft     \\ 
Edit Distance                     & lexical                      & Phrase           & Soft     \\ 
Edit Distance lemmatized          & lexical                      & Phrase           & Soft     \\ 
\end{tabular}
\caption{Overview of our labeling functions.}
\label{tab:overview_labeling_functions}
\end{table}

\subsection{Weak Supervision Aggregation}
During the development of our pipeline, we compared different techniques for aggregating the training signals of all labeling functions.
Therefore, we studied training \acp{HMM} and lightweight mathematical operations like the average, maximum, average-non-zero, and the normalized summation of all label scores from the labeling functions. According to our analysis, the HMM$_{V2}$ provides the most promising result which we used for the following steps in our pipeline.

\begin{landscape}
\begin{table}
\centering
\resizebox{1.25\textwidth}{!}{%
\begin{tabular}{llcccccccccc}

\textbf{Aggregation method}                      & \textbf{Final class} & \textbf{\begin{tabular}[c]{@{}l@{}}Average Number\\ of Justification Cues\end{tabular}} & \textbf{\begin{tabular}[c]{@{}l@{}}Average Number\\ of Tokens per \\ Justification Cue\end{tabular}} & \textbf{\begin{tabular}[c]{@{}l@{}}Average\\ Percentage of\\ Justification Cue \\ Tokens\end{tabular}} & \textbf{Avg} & \textbf{Dev} & \textbf{Median} & \textbf{Mode} & \textbf{Min} & \textbf{Max} & \textbf{Labeled Tokens} \\ \hline
\multirow{3}{*}{Average$_{non-hard}$}           & correct              & 0.11                                                                                    & 8.36                                                                                                 & 0.02                                                                                                   & 0.02         & 0.12         & 0.00            & 0.00          & 0.00         & 1.00         & 443                  \\ 
                                                 & partial correct      & 0.04                                                                                    & 6.29                                                                                                 & 0.01                                                                                                   & 0.01         & 0.07         & 0.00            & 0.00          & 0.00         & 1.00         & 88                   \\ 
                                                 & incorrect            & 0.00                                                                                    & 0.00                                                                                                 & 0.00                                                                                                   & 0.00         & 0.00         & 0.00            & 0.00          & 0.00         & 0.00         & 0                    \\ \hline
\multirow{3}{*}{HMM$_{V2}$}                      & correct              & 1.54                                                                                    & 13.85                                                                                                & 0.39                                                                                                   & 0.36         & 0.48         & 0.00            & 0.00          & 0.00         & 1.00         & 10,552               \\ 
                                                 & partial correct      & 1.25                                                                                    & 10.96                                                                                                & 0.36                                                                                                   & 0.28         & 0.45         & 0.00            & 0.00          & 0.00         & 1.00         & 4,594                \\ 
                                                 & incorrect            & 0.59                                                                                    & 9.28                                                                                                 & 0.17                                                                                                   & 0.21         & 0.41         & 0.00            & 0.00          & 0.00         & 1.00         & 464                  \\ \hline
\multirow{3}{*}{HMM$_{V1}$}                      & correct              & 2.06                                                                                    & 12.68                                                                                                & 0.52                                                                                                   & 0.44         & 0.50         & 0.00            & 0.00          & 0.00         & 1.00         & 12,937               \\ 
                                                 & partial correct      & 1.91                                                                                    & 11.00                                                                                                & 0.54                                                                                                   & 0.43         & 0.50         & 0.00            & 0.00          & 0.00         & 1.00         & 7,051                \\ 
                                                 & incorrect            & 1.06                                                                                    & 8.86                                                                                                 & 0.50                                                                                                   & 0.37         & 0.48         & 0.00            & 0.00          & 0.00         & 1.00         & 797                  \\ \hline
\multirow{3}{*}{Average$_{all}$}                 & correct              & 0.17                                                                                    & 3.51                                                                                                 & 0.02                                                                                                   & 0.01         & 0.10         & 0.00            & 0.00          & 0.00         & 1.00         & 288                  \\ 
                                                 & partial correct      & 0.10                                                                                    & 2.59                                                                                                 & 0.01                                                                                                   & 0.01         & 0.07         & 0.00            & 0.00          & 0.00         & 1.00         & 83                   \\ 
                                                 & incorrect            & 0.00                                                                                    & 0.00                                                                                                 & 0.00                                                                                                   & 0.00         & 0.00         & 0.00            & 0.00          & 0.00         & 0.00         & 0                    \\ \hline
\multirow{3}{*}{Max$_{non-hard}$}               & correct              & 1.00                                                                                    & 59.45                                                                                                & 1.00                                                                                                   & 1.00         & 0.00         & 1.00            & 1.00          & 1.00         & 1.00         & 29,485               \\ 
                                                 & partial correct      & 1.00                                                                                    & 48.65                                                                                                & 1.00                                                                                                   & 1.00         & 0.00         & 1.00            & 1.00          & 1.00         & 1.00         & 16,346               \\ 
                                                 & incorrect            & 1.00                                                                                    & 25.48                                                                                                & 1.00                                                                                                   & 1.00         & 0.00         & 1.00            & 1.00          & 1.00         & 1.00         & 2,166                \\ \hline
\multirow{3}{*}{Average-non-zero$_{non-hard}$} & correct              & 0.39                                                                                    & 5.73                                                                                                 & 0.05                                                                                                   & 0.04         & 0.19         & 0.00            & 0.00          & 0.00         & 1.00         & 1,118                \\ 
                                                 & partial correct      & 0.22                                                                                    & 4.87                                                                                                 & 0.05                                                                                                   & 0.02         & 0.15         & 0.00            & 0.00          & 0.00         & 1.00         & 365                  \\ 
                                                 & incorrect            & 0.07                                                                                    & 2.67                                                                                                 & 0.00                                                                                                   & 0.01         & 0.09         & 0.00            & 0.00          & 0.00         & 1.00         & 16                   \\ \hline
\multirow{3}{*}{Sum$_{non-hard}$}               & correct              & 1.44                                                                                    & 18.23                                                                                                & 0.38                                                                                                   & 0.44         & 0.50         & 0.00            & 0.00          & 0.00         & 1.00         & 13,038               \\ 
                                                 & partial correct      & 1.52                                                                                    & 15.15                                                                                                & 0.38                                                                                                   & 0.47         & 0.50         & 0.00            & 0.00          & 0.00         & 1.00         & 7,727                \\ 
                                                 & incorrect            & 0.74                                                                                    & 16.63                                                                                                & 0.28                                                                                                   & 0.48         & 0.50         & 0.00            & 0.00          & 0.00         & 1.00         & 1,048                \\ \hline
\multirow{3}{*}{Average-non-zero$_{all}$}       & correct              & 1.76                                                                                    & 4.65                                                                                                 & 0.17                                                                                                   & 0.14         & 0.34         & 0.00            & 0.00          & 0.00         & 1.00         & 4,070                \\ 
                                                 & partial correct      & 1.32                                                                                    & 3.87                                                                                                 & 0.19                                                                                                   & 0.11         & 0.31         & 0.00            & 0.00          & 0.00         & 1.00         & 1,717                \\ 
                                                 & incorrect            & 0.60                                                                                    & 2.75                                                                                                 & 0.15                                                                                                   & 0.06         & 0.25         & 0.00            & 0.00          & 0.00         & 1.00         & 140                  \\ \hline
\multirow{3}{*}{HMM$_{V3}$}                      & correct              & 1.23                                                                                    & 18.82                                                                                                & 0.34                                                                                                   & 0.39         & 0.49         & 0.00            & 0.00          & 0.00         & 1.00         & 11,516               \\ 
                                                 & partial correct      & 0.97                                                                                    & 19.03                                                                                                & 0.35                                                                                                   & 0.38         & 0.48         & 0.00            & 0.00          & 0.00         & 1.00         & 6,186                \\ 
                                                 & incorrect            & 0.52                                                                                    & 8.07                                                                                                 & 0.14                                                                                                   & 0.16         & 0.37         & 0.00            & 0.00          & 0.00         & 1.00         & 355                  \\ \hline
\multirow{3}{*}{Sum$_{all}$}                     & correct              & 3.66                                                                                    & 6.21                                                                                                 & 0.38                                                                                                   & 0.38         & 0.49         & 0.00            & 0.00          & 0.00         & 1.00         & 11,257               \\ 
                                                 & partial correct      & 2.90                                                                                    & 6.58                                                                                                 & 0.38                                                                                                   & 0.39         & 0.49         & 0.00            & 0.00          & 0.00         & 1.00         & 6,406                \\ 
                                                 & incorrect            & 2.07                                                                                    & 4.63                                                                                                 & 0.36                                                                                                   & 0.38         & 0.48         & 0.00            & 0.00          & 0.00         & 1.00         & 814                  \\ \hline
\multirow{3}{*}{Max$_{all}$}                     & correct              & 1.00                                                                                    & 59.45                                                                                                & 1.00                                                                                                   & 1.00         & 0.00         & 1.00            & 1.00          & 1.00         & 1.00         & 29,485               \\ 
                                                 & partial correct      & 1.00                                                                                    & 48.65                                                                                                & 1.00                                                                                                   & 1.00         & 0.00         & 1.00            & 1.00          & 1.00         & 1.00         & 16,346               \\ 
                                                 & incorrect            & 1.00                                                                                    & 25.48                                                                                                & 1.00                                                                                                   & 1.00         & 0.00         & 1.00            & 1.00          & 1.00         & 1.00         & 2,166                \\ 
\end{tabular}%
}
\caption{All results of the aggregation methods during our experiments in the weak supervision stage of the pipeline.}
\label{tab:results_all_aggregations}
\end{table}
\end{landscape}

\subsection{Scoring Vector Generation}
In our grading, the generation of the scoring vector is one of the most crucial steps. During development, we experimented with two different generation procedures.
As the first one implemented a hard matching, in which each justification cue matches exactly to one rubric element leading to sparse vectors.
The hard matching procedure assumes that the justification cue model correctly detects a justification cue's boundaries and achieves a high detection recall. 

As an alternative, we implemented fuzzy matching as the underlying assumption may not always hold. The fuzzy matching method matches all key elements from the scoring rubric to one justification cue. This generates a temporary scoring vector, which overrides the final scoring vector iteratively if a justification cue sets a new maximum value. As mentioned, this method allows us to not rely on the boundary detection of our justification cue model and generates dense scoring vectors. On the other hand, this methodology depends on the actual predicted scoring value, as it does not consider if a subset of justification cues is detected compared to the list of scoring rubric items. Furthermore, this implementation leads to scoring zero or non-zero vectors at all positions, which may affect their interpretability.

For our final configuration, we chose fuzzy matching.

\newpage

\subsection{Predictions}
We provide some exemplary predictions in Table \ref{tab:example_predictions} to compare our best-performing token classification and span prediction models extracted during our final analysis. 
\begin{table*}
\centering
\begin{tabular}{p{3cm}p{13cm}}
\textbf{ID}             & 1                                                                                                                                                                                                                                                                                                                                                                                  \\
\textbf{Question}       & What are the objectives of IPv6? Please state at least 4 objectives.                                                                                                                                                                                                                                                                                                               \\
\textbf{Student Answer} & 1. To support billions of end-systems 2. To reduce routing tables 3. To simplify protocol processing 4. To increase security 5. To support real time data traffic (quality of service) 6. To provide multicasting 7. To support mobility (roaming) 8. To be open for a change 9. To coexist with the existing protocol                                                             \\
\textbf{TC + DT}        & "To support billions of end-systems 2. To reduce routing tables 3. To simplify protocol processing 4. To increase security", "To support real time data traffic (quality of service) 6. To provide multicasting 7. To support mobility (roaming)", 'To be open for a change 9. To coexist with the existing protocol"                                                              \\
\textbf{SP + DT}        & "To support billions of end-systems 2.", "To reduce routing tables 3.", "To simplify protocol processing 4.", "To increase security 5.", "To support real time data traffic (quality of service) 6.", '', "To provide multicasting 7. To support mobility (roaming) 8.", "To support mobility (roaming) 8.", "To be open for a change 9.", "To coexist with the existing protocol" \\ \hline
\textbf{ID}             & 2                                                                                                                                                                                                                                                                                                                                                                                  \\
\textbf{Question}       & Frage 4: Ich muss die Sanitäranlagen besuchen, jedoch sind die Toiletten gesperrt. Wie gehe ich vor?                                                                                                                                                                                                                                                                               \\
\textbf{Student Answer} & per Fotos dokumentieren und schriftlich festhalten                                                                                                                                                                                                                                                                                                                                 \\
\textbf{TC + DT}        & "fest"                                                                                                                                                                                                                                                                                                                                                                             \\
\textbf{SP + DT}        & "Fotos", "Fotos dokumentieren und schriftlich festhalten"                                                                                                                                                                                                                                                                                                                          \\ \hline
\textbf{ID}             & 3                                                                                                                                                                                                                                                                                                                                                                                  \\
\textbf{Question}       & Frage 4: Ich muss die Sanitäranlagen besuchen, jedoch sind die Toiletten gesperrt. Wie gehe ich vor?                                                                                                                                                                                                                                                                               \\
\textbf{Student Answer} & Foto von außen machen, Info beifügen und keine andere Toilette die nicht zur Tankstelle gehört besuchen                                                                                                                                                                                                                                                                            \\
\textbf{TC + DT}        & "Foto von außen machen"                                                                                                                                                                                                                                                                                                                                                            \\
\textbf{SP + DT}        & "Foto von außen machen",  "Foto von außen machen"    \\                                 
\end{tabular}
\caption{Subset of predictions that allow comparison between our best-performing token classification (TC + DT$_{unseen}$) and span prediction model (SP + DT$_{bilingual}$).}
\label{tab:example_predictions}
\end{table*} 

\subsection{Question-specific Results} \label{appendix:question-specific-results}
To assess the question-specific performance, we interpreted the scores as a 9-class classification problem and generated the respective classification reports to gain deeper insights into our best-performing model. 
Therefore we rounded the scores within the values of 0 to 1 in steps of 0,125.
We provide all question-specific performances in Table \ref{tab:ASAS_question_results}.

\begin{table*}
\centering
\resizebox{0.7\textwidth}{!}{
\begin{tabular}{lcccc}
\textbf{Question ID} & \textbf{Acc.} & \textbf{Macro-F1} & \textbf{Weighted F1} & \textbf{Rubric Length} \\ \hline
1                    & 0.250         & 0.224             & 0.292                & 4                      \\
3                    & 0.241         & 0.133             & 0.278                & 3                      \\
4                    & 0.610         & 0.167             & 0.630                & 2                      \\
6                    & 0.019         & 0.017             & 0.036                & 2                      \\
7                    & 0.415         & 0.208             & 0.443                & 4                      \\
8                    & 0.103         & 0.037             & 0.175                & 2                      \\ 
5.12                 & 1.000         & 1.000             & 1.000                & 16                     \\
5.11                 & 0.957         & 0.489             & 0.935                & 1                      \\
4.3\_LM              & 0.700         & 0.275             & 0.741                & 2                      \\
12.2\_PE             & 0.500         & 0.242             & 0.485                & 5                      \\
10.2\_TC             & 0.182         & 0.052             & 0.165                & 7                      \\
2.1\_DLL\_v1.1       & 0.933         & 0.483             & 0.966                & 6                      \\
8.3\_MM              & 0.000         & 0.000             & 0.000                & 2                      \\
8.1\_MM              & 0.313         & 0.098             & 0.404                & 18                     \\
4.1\_LM\_v1.0        & 0.938         & 0.484             & 0.968                & 3                      \\
6.3                  & 0.375         & 0.154             & 0.363                & 5                      \\
2.4                  & 0.529         & 0.240             & 0.512                & 16                     \\
4.13                 & 0.385         & 0.180             & 0.321                & 2                      \\
10.1\_TC             & 0.583         & 0.156             & 0.583                & 11                     \\
6.1\_IPP             & 1.000         & 1.000             & 1.000                & 10                     \\
12.1\_PE             & 0.857         & 0.570             & 0.829                & 6                      \\
4.3                  & 0.100         & 0.047             & 0.047                & 2                      \\
10.3\_TC             & 0.500         & 0.400             & 0.500                & 7                      \\
2.3\_DLL\_v1.1       & 0.444         & 0.222             & 0.395                & 2                      \\
2.2\_DLL             & 0.917         & 0.478             & 0.877                & 2                      \\
12.3\_PE             & 0.071         & 0.050             & 0.089                & 2                      \\
6.2\_IPP             & 0.474         & 0.393             & 0.441                & 6                      \\
1.6                  & 0.792         & 0.181             & 0.792                & 4                      \\
8.2\_MM              & 0.077         & 0.026             & 0.014                & 3                      \\
6.3\_IPP             & 0.412         & 0.190             & 0.453                & 6                      \\
5.7                  & 0.357         & 0.228             & 0.379                & 7                      \\
4.2\_LM\_v1.0        & 0.786         & 0.229             & 0.851                & 2                     
\end{tabular}
}
\caption{Overview over the question-specific scoring performances as 9-class classification task from our best-performing model (SP + DT$_{bilingual}$).}
\label{tab:ASAS_question_results}
\end{table*}

\end{document}